\documentclass[letterpaper, 10 pt, journal, twoside]{IEEEtran}

\usepackage{amsmath,amssymb,amsfonts}
\usepackage{balance}
\usepackage{bm}
\usepackage{hyperref}
\usepackage{color}
\usepackage[table,xcdraw]{xcolor}
\usepackage{tabularx}
\usepackage{booktabs}
\usepackage{graphicx}
\usepackage{multirow}
\usepackage{tikz}
\usepackage{pgfplots}
\usepackage{siunitx}
\usetikzlibrary{bending, external}

\newcommand{\review}[1]{{\color{black} #1}}

\IEEEoverridecommandlockouts                             
\pgfplotsset{compat=1.16} 

\newcommand{\wfr}[0]{\ensuremath{\mathcal{W}}} %
\newcommand{\bfr}[0]{\ensuremath{\mathcal{B}}} %

\title{
Range, Endurance, and Optimal Speed Estimates for Multicopters}

\author{Leonard Bauersfeld$^{1}$ and Davide Scaramuzza$^{1}$
\thanks{Manuscript received: Sep. 10, 2021; Revised Dec. 8, 2021; Accepted Jan. 6, 2022.
This paper was recommended for publication by Editor Pauline Pounds upon evaluation of the Associate Editor and Reviewers' comments. This work was supported by the National Centre of Competence in Research (NCCR) Robotics through the Swiss National Science Foundation (SNSF) and the European Union’s Horizon 2020 Research and Innovation Programme under grant agreement No. 871479 (AERIAL-CORE) and the European Research Council (ERC) under grant agreement No. 864042 (AGILEFLIGHT).}
\thanks{$^1$ The authors are with the Robotics and Perception Group, Department of Informatics, University of Zurich, and Department of Neuroinformatics, University of Zurich and ETH Zurich, Switzerland (\protect\url{http://rpg.ifi.uzh.ch})  {\tt\footnotesize bauersfeld@ifi.uzh.ch}. 
}
\thanks{Digital Object Identifier (DOI): see top of this page.}
}

\begin{document}

\maketitle

\begin{abstract}
Multicopters are among the most versatile mobile robots. Their applications range from inspection and mapping tasks to providing vital reconnaissance in disaster zones and to package delivery. The range, endurance, and speed a multirotor vehicle can achieve while performing its task is a decisive factor not only for vehicle design and mission planning, but also for policy makers deciding on the rules and regulations for aerial robots.
To the best of the authors' knowledge, this work proposes the first approach to estimate the range, endurance, and optimal flight speed for a wide variety of multicopters. This advance is made possible by combining a state-of-the-art first-principles aerodynamic multicopter model based on blade-element-momentum theory with an electric-motor model and a graybox battery model. This model predicts the cell voltage with only 1.3\% relative error (43.1\,mV), even if the battery is subjected to non-constant discharge rates. Our approach is validated with real-world experiments on a test bench as well as with flights at speeds up to 65\,km/h in one of the world's largest motion-capture systems.
We also present an accurate pen-and-paper algorithm to estimate the range, endurance and optimal speed of multicopters to help future researchers build drones with maximal range and endurance, ensuring that future multirotor vehicles are even more versatile.
\end{abstract}
\vspace*{-2pt}
\begin{IEEEkeywords}
Aerial Systems: Mechanics and Control, Motion and Path Planning, Optimization and Optimal Control
\end{IEEEkeywords}
\setlength{\abovedisplayskip}{6pt}
\setlength{\belowdisplayskip}{6pt}
\setlength{\abovedisplayshortskip}{4pt}
\setlength{\belowdisplayshortskip}{4pt}

\vspace*{-3pt}
\section{Introduction}
\vspace*{-1pt}

\IEEEPARstart{I}{n} the recent years, autonomous aerial multirotor vehicles (also known as multicopters) have been adopted for a wide variety of tasks~\cite{2014:DAndrea, 2017:Karydis, 2018:Mohta}. International, multi-million-dollar projects such as AgileFlight~\cite{AgileFlight}, Aerial-Core (autonomous power line inspection)
, DARPA FLA (fast lightweight autonomy)
and AlphaPilot~\cite{2020:Foehn}  (autonomous drone racing) each helped pushing the frontiers of research in their respective field. However, one problem common to all applications of multirotor vehicles is often not considered: their range and endurance is very limited compared to other mobile robots since they have a much higher energy consumption than ground vehicles or fixed-wing aircraft. 

\begin{figure}
    \centering
    \vspace*{1pt}
    \includegraphics[]{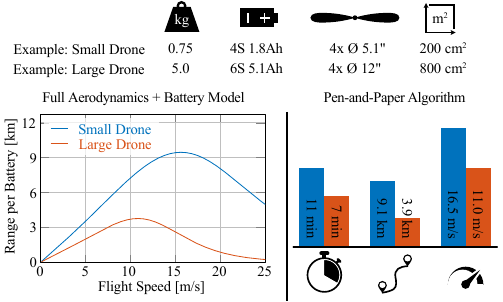}
    \vspace*{-8pt}
    \caption{\review{The approach presented in this work can be used to calculate accurate range, endurance, and optimal speed estimates for general multicopters. Depending on the accuracy requirements, either a state-of-the-art blade-element-momentum aerodynamics simulation can be combined with an accurate motor and battery model, or alternatively a simple, yet precise pen-and-paper algorithm can be employed to calculate the performance estimates. For the latter, the only information required is the mass, battery type, propeller size, and average surface area of the multicopter.}}
    \vspace*{-15pt}
    \label{fig:overview}
\end{figure}

This paper proposes an approach to obtain accurate range, endurance, and optimal speed estimates for multicopters. Having access to this information (see Fig.~\ref{fig:overview}) helps researchers and companies alike to optimize their mechanical design and mission planning towards meeting given specifications such as required flight times or operating radii. Knowledge of feasible flight distances and speeds also enables policy-makers to make informed decisions on the regulations for multicopter use. Lastly, understanding the tradeoffs between range, mass, speed, and agility is important when assessing the suitability of a multicopter for a new task.

Estimating the range and endurance of a multicopter requires an accurate model of the vehicle's power consumption. This is particularly difficult because the instantaneous power draw is influenced by the airflow around the vehicle~\cite{1995:Prouty}, by the rotor speeds of the individual motors, and by the particular motor-propeller combination~\cite{2020:Biczyski}. Furthermore, a battery model that still accurately holds when the commonly used LiPo batteries (lithium-polymer) are discharged at very high rates is required.

Existing approaches for range and endurance estimates often focus on hover-endurance~\cite{2019:Hnidka, 2019:Lussier}, where the complex aerodynamic effects of multicopter flight can be neglected. Works concerned with estimating the maximum flight range use simplistic multicopter models for forward flight~\cite{2018:Hwang, 2019:Godbole, 2017:Cieslewski}. Such models neglect key aerodynamic effects experienced by all rotary wing aircraft: as the vehicle flies forward, the dynamic lift experienced by the propellers yields a reduction in the multicopter's power consumption. Furthermore, linear rotor drag (induced drag) is not considered albeit significantly contributing to the overall drag~\cite{2014:Ducard}. 

The above mentioned works either assume that the battery is an ideal energy storage or use the Peukert model~\cite{1897:Peukert} to calculate the battery capacity. However, this model is only accurate for the low to medium discharge rates~\cite{2020:Galushkin} typically encountered in ground vehicles or fixed-wing aircraft but not well-suited for the very high power demand of multicopters.

\vspace*{-12pt}
\subsection*{Contribution}
\vspace*{-2pt}

To the best of the authors' knowledge, this work presents the first approach to accurately determine range, endurance, and optimal flight speed estimates for general multicopters. This advance is made possible by combining a state-of-the-art first-principles aerodynamic multicopter model based on blade-element-momentum theory (BEM)~\cite{2021:Bauersfeld} with both a graybox motor model and battery model, each identified from various motors and batteries. The battery model is highly accurate when compared with experimental data in both a constant discharge-rate and a variable discharge-rate setting, yielding an average RMSE of \SI{43.1}{\milli\volt} per cell (1.3\,\%). The BEM model is validated against real-world flight data at speeds up to \SI{65}{\kilo\meter\per\hour} recorded in \review{a very large optical tracking volume} \footnote{\url{https://www.youtube.com/watch?v=EV4ACi5ZO2k}}. It achieves an average thrust prediction error of \SI{0.91}{\newton} and an mean power prediction error of \SI{33}{\watt} (2.7\,\% of peak power).

Based on the proposed method, we present a pen-and-paper algorithm to calculate the range, endurance, and optimal flight speed of a multirotor vehicle based on its propeller diameter, its battery capacity, its size, and its mass. This method has also been used to generate Fig.~\ref{fig:overview}.

\vspace*{-3pt}
\section{Related Work}
\vspace*{-1pt}
A general overview of the field of energetics in robotics flight is presented in~\cite{2017:Karydis}. Although the survey paper touches this work's topic only briefly, it outlines some of the key problems encountered when estimating range, endurance, and optimal flight speed of multirotor aerial vehicles: accurately modeling the aerodynamics and the power source.

First, the work related to modeling the aerodynamic forces and torques acting on a multicopter is summarized. When modeling such aerodynamic wrenches, it is commonly assumed that each propeller produces a thrust force and an axial torque proportional to the square of its rotational speed~\cite{2012:Mahony, 2016:Furrer, 2018:Shah, 2020:Song}. 
This \emph{quadratic model} holds very well for multicopters in hover flight, but becomes increasingly inaccurate as the vehicle flies faster because it neglects important aerodynamic effects. The most prominent unmodeled effect is the induced propeller drag (linear drag), which can be incorporated into the model by adding a velocity dependent drag term~\cite{2016:Furrer, 2014:Ducard}. However, more involved aerodynamic effects such as dynamic lift cannot be accurately accounted for. Dynamic lift is a phenomenon encountered by all rotary-wing aircraft where the thrust of the propeller increases when the in-plane (i.e. in the propeller blade plane) airspeed increases~\cite{2017:Karydis}. Due to the modeling inaccuracies in forward flight, the quadratic model is not well suited for range estimates and its use for endurance estimation should be limited to hover endurance. To overcome the limitations of the quadratic model, blade-element-momentum (BEM) theory can be used. BEM theory is known to accurately model the aerodynamic forces and torques acting on a pro-peller across a wide range of airspeeds~\cite{1995:Prouty, 2017:Gill, 2007:Hoffmann, 2013:Khan}. 

Next to the well-established first-principles models, a recent line of work on machine-learned multicopter models has emerged~\cite{2019:Sun, 2021:Torrente, 2019:Shi, 2018:Mohajerin}. Despite being accurate, they are not well suited for range, endurance, and speed estimation of general multicopters because they do not predict the power consumption and only apply to the exact vehicle they have been trained on. 

The earliest work on battery modeling dates back to the late 19th century, when Peukert studied how the capacity of lead-acid batteries depends on the discharge current~\cite{1897:Peukert}. Due to its simplicity, the \emph{Peukert Model} has since become the standard approach to model the \emph{effective capacity} under load. It has also been shown to hold for LiPo batteries at medium discharge rates~\cite{2020:Galushkin, 2020:Gong}. Generalizations to medium discharge (around 1\,C) rates exist as well~\cite{2015:Galushkin, 2020:Galushkin}. 

To overcome this limitation and mainly to directly model the cell voltage of a battery, a graybox battery model based on a Thevenin \emph{equivalent circuit} can be used. Depending on the fidelity of the model, it includes one resistor combined with zero, one (one time constant, OTC) or two (two time constants, TTC) capacitive networks. A review of the common OTC and TTC models is presented in~\cite{2016:Zhang}. The OTC model is widely used because of its well-established accuracy~\cite{2020:Zhang}. A TTC model only shows improved precision in cases where the battery dynamics need to be accurately captured at very short timescales~\cite{2012:Rahmoun}.
Much more elaborate battery models based on molecular dynamics simulations exist~\cite{2018:Muralidharan, 2019:Aoyagi} but are unsuited for the task at hand because of being highly specific to exactly one battery type.

Only very few studies try to estimate the range, endurance or optimal flight speed by combining a multicopter aerodynamics model with a battery model~\cite{2017:Karydis, 2019:Hnidka, 2019:Lussier, 2020:Biczyski, 2019:Godbole, 2018:Hwang, 2017:Cieslewski}. The first group of works focuses on the hover endurance of multirotor aerial vehicles. In~\cite{2019:Hnidka} BEM theory is used to calculate the required power to hover and it is combined with a purely measured battery model. In~\cite{2019:Lussier} BEM is also employed but the focus lies on alternative power sources such as hydrogen cells. Albeit mainly focused on motor and propeller selection for UAV's,~\cite{2020:Biczyski} presents a hover endurance estimate. A quadratic model for the aerodynamics and a Peukert model for the battery are used. The second group of works additionally focuses on calculating the range and optimal flight speed. In~\cite{2019:Godbole} an ideal battery model is used together with a quadratic propeller model augmented with quadratic body drag. The approach neglects induced propeller drag, dynamic lift and assumes an ideal battery with no Peukert effect. A very similar approach is followed in~\cite{2017:Cieslewski}, but induced propeller drag is additionally considered. Both works present results only in simulation. A more thorough approach is presented in~\cite{2018:Hwang}. They use a momentum-theory model to calculate the required power during forward flight and combine this model with a Peukert battery model. However, the dominant induced drag is neglected and no general range, endurance or flight speed estimates for other vehicles than the one studied are provided.

The approach taken in this work is inspired by the survey presented in~\cite{2017:Karydis} and influenced by~\cite{2011:Traub} where range and endurance estimates for battery-powered fixed-wing aircraft are presented. We use a state-of-the-art BEM model~\cite{2021:Bauersfeld} together with a body-drag model to calculate the power a multicopter requires to fly at a given speed. The battery dynamics are modeled using a one-time-constant model (OTC)~\cite{2012:Rahmoun}. To improve the accuracy of the range and endurance estimates even further, a graybox model for brushless motor efficiency informed by~\cite{2020:Biczyski} is developed.

\vspace*{-4pt}
\section{Aerodynamics Simulator}
\vspace*{-1pt}
This section briefly explains the multicopter simulator. The dynamics of a multirotor can be written as 
\begin{align}
\small
\label{eq:3d_quad_dynamics}
\dot{\bm{x}} =
\begin{bmatrix}
\dot{\bm{p}}_{\wfr\bfr} \\  
\dot{\bm{q}}_{\wfr\bfr} \\
\dot{\bm{v}}_{\wfr} \\
\dot{\boldsymbol\omega}_\bfr
\end{bmatrix} = 
\begin{bmatrix}
\bm{v}_\wfr \\  
\bm{q}_{\wfr\bfr} \cdot \begin{bmatrix}
0 \\ \bm{\omega}_\bfr/2\end{bmatrix} \\
\frac{1}{m} \Big(\bm{q}_{\wfr\bfr} \odot (\bm{f}_\text{prop} + \bm{f}_\text{body})\Big)+\bm{g}_\wfr  \\
\bm{J}^{-1}\big( \boldsymbol{\tau}_\text{prop}  - \boldsymbol\omega_\bfr \times \bm{J}\boldsymbol\omega_\bfr \big)
\end{bmatrix} \; ,
\end{align}
where $\bm{p}_{\wfr\bfr}$, $\bm{q}_{\wfr\bfr}$, $\bm{v}_{\wfr}$, and $\boldsymbol\omega_\bfr$ denote the position, attitude quaternion, inertial velocity, and bodyrates of the multicopter, respectively. The matrix $\bm J$ is the multicopter's inertia and $\bm{g}_\wfr$ denotes the gravity vector. The propeller force $\bm f_\text{prop}$, body drag force $\bm f_\text{body}$ and propeller torque $\boldsymbol{\tau}_\text{prop}$ acting on the vehicle are given as
\vspace*{-5pt}
\begin{align}
   \bm{f}_\text{prop} &= \sum\nolimits_i \bm{f}_i\,,\qquad 
   \boldsymbol{\tau}_\text{prop} = \sum\nolimits_i \boldsymbol{\tau}_i + \bm{r}_{\text{P},i} \times \bm{f}_i \,, \label{eq:torques}\\
   \bm{f}_\text{body} &= -0.5 c_\text{body} \rho \, |\bm A \cdot \review{\bm{v}_{\bfr, \text{rel}}|  \, \bm{v}_{\bfr, \text{rel}}}\,, \label{eq:forces}
\end{align}

\noindent \review{with $v_{\bfr, \text{rel}}$ the relative airspeed in body frame,} $c_\text{body}$ the drag coefficient of the body, $\bm A$ the vector of reference surface areas, $\rho$ the air density, $\bm{r}_{\text{P},i}$ the location of propeller $i$ expressed in the body frame and $\bm{f}_i$, $\boldsymbol{\tau}_i$ the forces and torques generated by the $i$-th propeller.

\vspace*{-9pt}
\subsection{BEM Model}
\vspace*{-2pt}
To accurately model the forces $\bm f_i$ and torques $\boldsymbol\tau_i$ in (\ref{eq:forces}), (\ref{eq:torques}), a blade-element-momentum theory (BEM) model is used. BEM models the physical process that generates the thrust force and the axial drag torque, i.e. it models how each infinitesimal blade element of a propeller creates a force and torque that are then integrated to yield the overall thrust and rotor drag~\cite{1995:Prouty, 2017:Gill, 2007:Hoffmann}. 

In order to model the lift and drag of each blade element, the airflow around it needs to be determined. The free-flow velocity is directly known from the multicopter's ego-motion. However, the propeller itself accelerates the air downwards. This  \emph{induced velocity} $v_i$ can be calculated by combining blade-element theory and momentum theory, hence the name BEM. Momentum theory is a simple theory that calculates the thrust $T$ based on a momentum balance inside a flow tube across the propeller.
The induced velocity $v_i$ is found such that the thrust calculated using momentum theory equals the sum of the infinitesimal lift forces over all blade elements (see~\cite{1995:Prouty, 2017:Gill, 2007:Hoffmann} for details). The BEM model used in this work also accounts for \review{oblique inflow \cite{2015:Khan, 2017:Theys}} and even more complex effects like blade elasticity and the resulting blade flapping. An empirical model~\cite{2007:Hoffmann} to compute the induced velocity when the multicopter is in vortex-ring-state~\cite{1995:Prouty} is also implemented. For a more in-depth treatment of the topic \review{and details about the model used}, the reader is referred to our previous work~\cite{2021:Bauersfeld}. 

In this work it is also shown that the BEM model is very accurate in predicting the forces and torques acting on a multicopter. On a set of very aggressive test trajectories with speeds up to \SI{65}{\kilo\meter\per\hour}, it achieves an RMS error when predicting the vehicle's thrust of less than \SI{0.91}{\newton}. 

\vspace*{-11pt}
\subsection{Hover Flight}
\vspace*{-3pt}
The complexity of the BEM model is necessary to model the forces and torques acting on the multicopter throughout its entire performance envelope. However, when the multicopter is in hover, momentum theory alone can be used to calculate the induced velocity and mechanical power:
\begin{equation}
    v_{i,h} = \sqrt{\frac{T_h}{2 \rho A_\text{prop}}} = \sqrt{\frac{\vphantom{T}m g}{2 \rho \pi r_\text{prop}^2 N_r}}\,,
    \label{eq:v_ind_h}
\end{equation}
where $T_h$ is the thrust of each propeller required to hover, $m$ is the mass of the multicopter and $N_r$ its number of rotors with radius $r_\text{prop}$. Based on this, the mechanical hover power of a multicopter can be calculated:
\begin{equation}
    P_{h} = \frac{N_r T_h v_{i,h}}{\eta_P} = \frac{(mg)^{3/2}}{\eta_P \sqrt{2 \rho \pi N_r} r_\text{prop}}
    \label{eq:hover_power}
\end{equation}
where $\eta_P$ is the figure of merit (propeller efficiency). Typical propellers achieve a figure of merit between 0.5~\cite{2012:Deters} and 0.7~\cite{2014:Sforza}. A value of $\eta_P = 0.6$ is used subsequently. 
The geometric pitch of the propeller is not taken into account by momentum theory in (\ref{eq:hover_power}). This is confirmed by experimental data indicating that the power to produce a given thrust at hover does not depend on the propeller pitch.

\vspace*{-6pt}
\section{Motor Model}
\vspace*{-1pt}
The BEM model outlined above computes accurate axial torque predictions $Q$ for the multicopter's propellers. The motor model presented in this section is then used to calculate the power consumption of each motor $P_\text{mot}$ as
\begin{equation}
    P_\text{mot}(t) = \frac{Q(t) \cdot \Omega(t)}{\eta_M(\Omega)}\,.
    \label{eq:motor_model}
\end{equation}
It is assumed that the motor efficiency $\eta_M$ is only a function of the rotational speed $\Omega$.

\vspace*{-11pt}
\subsection{Derivation of Efficiency Model}
\vspace*{-2pt}
Existing work on brushless motors~\cite{2020:Biczyski} and experimental data show that the motor efficiency depends on the motor speed. Based on physical insights, a more accurate model is developed.

The total power consumption is assumed to be the sum of a mechanical power and a electrical loss term $P_\text{loss}$:
\begin{equation}
    P_\text{mot} = P_\text{mech} + P_\text{loss} = \Omega \left(Q + m_0\right) + P_\text{loss}
    \label{eq:ptot}
\end{equation}
where $Q$ is the aerodynamic drag torque of the propeller and $ m_0$ is a sliding friction coefficient. A straightforward choice for $P_\text{loss}$ would be to account for electric losses due to the internal resistance of the motor:
\begin{equation}
    P_\text{loss} = R_\text{i} I_\text{mot}^2 = R_\text{i} \left(\frac{P_\text{mot}}{U_\text{mot}}\right)^2 \approx R_\text{i} \left(\frac{c_d \Omega^3}{U_\text{mot}}\right)^2\,.
    \label{eq:ploss}
\end{equation}
The last step assumes that the dominant mechanical drag torque is due to the propeller and can be approximated as $Q = c_d \Omega^2$, where $c_d$ is the drag coefficient of the propeller.
Under the assumption that the motor voltage is constant, (\ref{eq:ptot}) and (\ref{eq:ploss}) can be combined. Together with (\ref{eq:motor_model}), this yields a motor model of the form
\begin{equation}
    \eta_M(\Omega) = \frac{c_d \Omega^3}{m_0 \Omega + m_1 \Omega^3 + m_2 \Omega^6}
    \label{eq:eta}
\end{equation}
where the coefficients $c_d$, $m_0$, $m_1$, $m_2$ of the lumped parameter model depend on the motor-propeller combination.

\vspace*{-9pt}
\subsection{Experimental Validation}
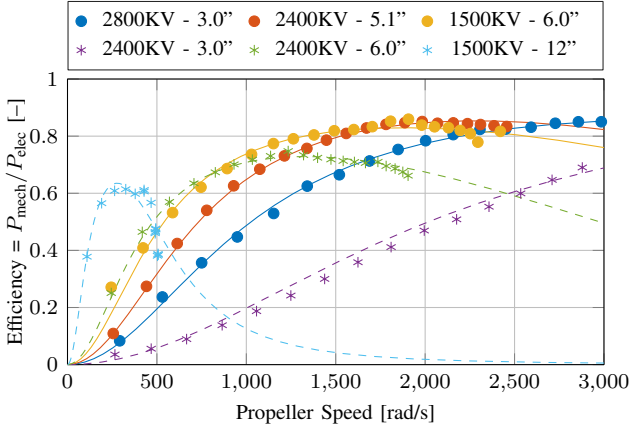
\begin{figure}
     \centering 
    \vspace*{2pt}
    \tikzstyle{every node}=[font=\footnotesize]
\definecolor{mycolor1}{rgb}{0.00000,0.44700,0.74100}%
\definecolor{mycolor2}{rgb}{0.85000,0.32500,0.09800}%
\definecolor{mycolor3}{rgb}{0.92900,0.69400,0.12500}%
\definecolor{mycolor4}{rgb}{0.49400,0.18400,0.55600}%
\definecolor{mycolor5}{rgb}{0.46600,0.67400,0.18800}%
\definecolor{mycolor6}{rgb}{0.30100,0.74500,0.93300}%
\begin{tikzpicture}

\begin{axis}[%
width=7.132cm,
height=3.8cm,
at={(0cm,0cm)},
scale only axis,
xmin=0.0000,
xmax=3000.0000,
xlabel={Propeller Speed [rad/s]},
ymin=0.0000,
ymax=1.0000,
ylabel={Efficiency = $P_{\text{mech}}/P_{\text{elec}}$ [\textendash]},
axis background/.style={fill=white},
xmajorgrids,
ymajorgrids,
legend style={legend cell align=left, align=left, draw=white!15!black},
legend columns=3,
xlabel shift = -2pt,
ylabel shift = -6pt,
line join = round,
scale only axis=true,
legend style={at={(axis cs:0,1.03)}, anchor=south west,  minimum width=2.16cm}
]
\addplot [color=mycolor1, forget plot]
  table[row sep=crcr]{%
1.0000	0.0000\\
49.0000	0.0023\\
97.0000	0.0091\\
147.0000	0.0206\\
199.0000	0.0371\\
254.0000	0.0590\\
315.0000	0.0879\\
384.0000	0.1251\\
470.0000	0.1762\\
628.0000	0.2756\\
768.0000	0.3616\\
872.0000	0.4210\\
969.0000	0.4721\\
1064.0000	0.5176\\
1160.0000	0.5591\\
1259.0000	0.5974\\
1362.0000	0.6328\\
1470.0000	0.6655\\
1585.0000	0.6958\\
1708.0000	0.7237\\
1841.0000	0.7494\\
1985.0000	0.7728\\
2143.0000	0.7938\\
2316.0000	0.8124\\
2506.0000	0.8282\\
2715.0000	0.8412\\
2946.0000	0.8509\\
3000.0000	0.8526\\
};
\addplot [color=mycolor1, only marks, mark=*, mark options={solid, fill=mycolor1, mycolor1}]
  table[row sep=crcr]{%
291.8976	0.0834\\
530.6053	0.2369\\
750.2253	0.3565\\
949.6198	0.4471\\
1152.8220	0.5293\\
1340.6823	0.6250\\
1520.1119	0.6648\\
1689.2169	0.7131\\
1848.8513	0.7533\\
2002.6674	0.7839\\
2156.9331	0.8053\\
2306.3765	0.8236\\
2450.7360	0.8247\\
2592.4856	0.8313\\
2727.9457	0.8451\\
2857.0127	0.8503\\
2985.0769	0.8503\\
};
\addlegendentry{2800KV - 3.0''}

\addplot [color=mycolor2, forget plot]
  table[row sep=crcr]{%
1.0000	0.0000\\
36.0000	0.0024\\
71.0000	0.0093\\
107.0000	0.0209\\
144.0000	0.0372\\
184.0000	0.0594\\
228.0000	0.0884\\
278.0000	0.1260\\
340.0000	0.1773\\
446.0000	0.2705\\
554.0000	0.3639\\
630.0000	0.4251\\
700.0000	0.4770\\
768.0000	0.5231\\
837.0000	0.5653\\
908.0000	0.6043\\
981.0000	0.6399\\
1058.0000	0.6730\\
1139.0000	0.7033\\
1226.0000	0.7313\\
1319.0000	0.7567\\
1419.0000	0.7796\\
1527.0000	0.7997\\
1644.0000	0.8170\\
1771.0000	0.8313\\
1909.0000	0.8423\\
2059.0000	0.8498\\
2221.0000	0.8534\\
2396.0000	0.8528\\
2585.0000	0.8477\\
2789.0000	0.8376\\
3000.0000	0.8230\\
};
\addplot [color=mycolor2, only marks, mark=*, mark options={solid, fill=mycolor2, mycolor2}]
  table[row sep=crcr]{%
255.8463	0.1090\\
441.5652	0.2741\\
613.4731	0.4240\\
778.6662	0.5403\\
929.3274	0.6261\\
1075.2758	0.6842\\
1213.0649	0.7309\\
1338.5843	0.7570\\
1448.0380	0.7860\\
1558.5727	0.8095\\
1671.9508	0.8285\\
1781.2252	0.8412\\
1886.3249	0.8471\\
1983.6317	0.8511\\
2069.1134	0.8451\\
2157.4117	0.8464\\
2236.9161	0.8437\\
2309.5352	0.8406\\
2385.2867	0.8363\\
2456.3746	0.8338\\
};
\addlegendentry{2400KV - 5.1''}

\addplot [color=mycolor3, forget plot]
  table[row sep=crcr]{%
1.0000	0.0000\\
25.0000	0.0023\\
50.0000	0.0092\\
76.0000	0.0210\\
103.0000	0.0379\\
132.0000	0.0606\\
164.0000	0.0904\\
201.0000	0.1294\\
249.0000	0.1848\\
405.0000	0.3681\\
456.0000	0.4217\\
505.0000	0.4688\\
554.0000	0.5114\\
604.0000	0.5504\\
656.0000	0.5864\\
710.0000	0.6194\\
768.0000	0.6503\\
830.0000	0.6788\\
897.0000	0.7051\\
970.0000	0.7292\\
1050.0000	0.7511\\
1138.0000	0.7707\\
1236.0000	0.7879\\
1345.0000	0.8026\\
1467.0000	0.8144\\
1603.0000	0.8231\\
1754.0000	0.8281\\
1920.0000	0.8292\\
2102.0000	0.8258\\
2300.0000	0.8177\\
2513.0000	0.8044\\
2744.0000	0.7855\\
2996.0000	0.7604\\
3000.0000	0.7600\\
};
\addplot [color=mycolor3, only marks, mark=*, mark options={solid, fill=mycolor3, mycolor3}]
  table[row sep=crcr]{%
2420.5011	0.8168\\
244.5220	0.2709\\
422.3666	0.4087\\
587.6023	0.5322\\
746.4488	0.6212\\
892.5088	0.6864\\
1029.0976	0.7361\\
1151.5719	0.7739\\
1263.2337	0.7908\\
1378.3737	0.8042\\
1491.7694	0.8188\\
1601.0339	0.8230\\
1704.5247	0.8332\\
1806.5296	0.8523\\
1906.7656	0.8594\\
1981.0325	0.8386\\
2052.5753	0.8328\\
2130.3905	0.8295\\
2195.8294	0.8200\\
2252.7637	0.8088\\
2294.6540	0.7790\\
};
\addlegendentry{1500KV - 6.0''}

\addplot [color=mycolor4, dashed, forget plot]
  table[row sep=crcr]{%
1.0000	0.0000\\
96.0000	0.0023\\
192.0000	0.0090\\
291.0000	0.0204\\
394.0000	0.0368\\
503.0000	0.0585\\
622.0000	0.0868\\
758.0000	0.1237\\
925.0000	0.1738\\
1195.0000	0.2598\\
1520.0000	0.3622\\
1732.0000	0.4244\\
1927.0000	0.4772\\
2117.0000	0.5242\\
2309.0000	0.5672\\
2506.0000	0.6069\\
2712.0000	0.6439\\
2929.0000	0.6783\\
3000.0000	0.6887\\
};
\addplot [color=mycolor4, only marks, mark=asterisk, mark options={solid, fill=mycolor4, mycolor4}]
  table[row sep=crcr]{%
265.1448	0.0355\\
468.0469	0.0558\\
665.9655	0.0894\\
864.8678	0.1380\\
1057.0195	0.1873\\
1249.0193	0.2419\\
1437.9542	0.3001\\
1624.3446	0.3585\\
1809.5593	0.4115\\
1994.9902	0.4694\\
2175.9654	0.5087\\
2357.1282	0.5529\\
2534.4030	0.5995\\
2709.7582	0.6469\\
2878.8848	0.6902\\
};
\addlegendentry{2400KV - 3.0''}

\addplot [color=mycolor5, dashed, forget plot]
  table[row sep=crcr]{%
1.0000	0.0000\\
21.0000	0.0024\\
41.0000	0.0091\\
62.0000	0.0206\\
84.0000	0.0371\\
108.0000	0.0596\\
135.0000	0.0895\\
167.0000	0.1296\\
214.0000	0.1935\\
298.0000	0.3079\\
342.0000	0.3629\\
383.0000	0.4097\\
423.0000	0.4508\\
464.0000	0.4884\\
506.0000	0.5225\\
550.0000	0.5537\\
597.0000	0.5826\\
647.0000	0.6089\\
702.0000	0.6332\\
761.0000	0.6549\\
826.0000	0.6742\\
898.0000	0.6911\\
977.0000	0.7051\\
1065.0000	0.7163\\
1162.0000	0.7240\\
1269.0000	0.7281\\
1387.0000	0.7280\\
1515.0000	0.7235\\
1654.0000	0.7141\\
1806.0000	0.6992\\
1973.0000	0.6784\\
2160.0000	0.6505\\
2380.0000	0.6131\\
2673.0000	0.5585\\
3000.0000	0.4953\\
};
\addplot [color=mycolor5, only marks, mark=asterisk, mark options={solid, fill=mycolor5, mycolor5}]
  table[row sep=crcr]{%
244.6686	0.2507\\
418.2119	0.4647\\
570.1068	0.5697\\
707.2160	0.6347\\
826.6180	0.6732\\
931.0810	0.7012\\
1033.7159	0.7178\\
1135.5583	0.7306\\
1234.0882	0.7477\\
1319.9802	0.7338\\
1398.3706	0.7244\\
1473.3011	0.7184\\
1541.6648	0.7132\\
1604.5880	0.7095\\
1666.2516	0.7061\\
1735.6843	0.7100\\
1781.8991	0.6954\\
1831.9590	0.6864\\
1875.1280	0.6749\\
1904.5642	0.6616\\
};
\addlegendentry{2400KV - 6.0''}

\addplot [color=mycolor6, dashed, forget plot]
  table[row sep=crcr]{%
1.0000	0.0001\\
7.0000	0.0029\\
13.0000	0.0100\\
19.0000	0.0212\\
26.0000	0.0387\\
33.0000	0.0607\\
41.0000	0.0900\\
51.0000	0.1314\\
66.0000	0.1987\\
88.0000	0.2970\\
101.0000	0.3507\\
113.0000	0.3958\\
125.0000	0.4363\\
137.0000	0.4722\\
149.0000	0.5035\\
162.0000	0.5328\\
175.0000	0.5575\\
189.0000	0.5795\\
204.0000	0.5983\\
219.0000	0.6128\\
235.0000	0.6238\\
252.0000	0.6313\\
271.0000	0.6350\\
291.0000	0.6343\\
312.0000	0.6292\\
335.0000	0.6191\\
361.0000	0.6031\\
390.0000	0.5807\\
425.0000	0.5489\\
472.0000	0.5013\\
604.0000	0.3652\\
654.0000	0.3195\\
701.0000	0.2810\\
748.0000	0.2468\\
797.0000	0.2157\\
849.0000	0.1872\\
904.0000	0.1615\\
964.0000	0.1381\\
1030.0000	0.1168\\
1104.0000	0.0975\\
1188.0000	0.0802\\
1285.0000	0.0647\\
1400.0000	0.0509\\
1539.0000	0.0390\\
1712.0000	0.0287\\
1936.0000	0.0200\\
2240.0000	0.0131\\
2684.0000	0.0076\\
3000.0000	0.0055\\
};
\addplot [color=mycolor6, only marks, mark=asterisk, mark options={solid, fill=mycolor6, mycolor6}]
  table[row sep=crcr]{%
109.0589	0.3779\\
190.7761	0.5652\\
263.4418	0.6086\\
325.7400	0.6142\\
378.6547	0.5979\\
428.0902	0.6063\\
428.2069	0.6115\\
467.5665	0.5670\\
493.4836	0.4766\\
493.4743	0.4734\\
490.8238	0.4646\\
490.3437	0.4600\\
507.4713	0.3845\\
506.2975	0.3875\\
503.1040	0.3811\\
};
\addlegendentry{1500KV - 12''}

\end{axis}
\end{tikzpicture}%
\tikzstyle{every node}=[font=\normalsize]
    \vspace*{-10pt}
    \caption{Efficiency of six different motor-propeller combinations plotted over the entire operating range of each motor. The lines represent the fitted motor model (\ref{eq:eta}). Solid lines (circle marks) represent a motor-propeller combination recommended by the manufacturer, whereas the dashed lines (star marks) show mismatched pairings.}
    \label{fig:motor_efficiency}
    \vspace*{-15pt}
\end{figure}

\vspace*{-1pt}
To validate the model, 44 different motor-propeller combinations have been measured on a thrust-test stand. The recorded data contains motor speeds, generated thrust, aerodynamic drag torque, and power consumption. Fig.~\ref{fig:motor_efficiency} exemplarily shows measured efficiencies along with the fitted models (\ref{eq:eta}) for six different motor-propeller pairings. 

Motor-propeller pairings following the manufacturers' recommendations (solid lines, circle marks) follow a very similar shape. They achieve 70-85\,\% efficiency for a wide range of operating conditions. If the propeller is too small for the motor (2400KV - 3.0''), the mechanical friction lowers the overall efficiency unless the motor is operated at very high speeds. If the propeller is too large (2400KV - 6.0'', 1500KV - 12''), the efficiency drastically decreases at higher propeller speeds because the motor \emph{stalls} (is unable to achieve the commanded speed) and gets very hot.

Because all recommended motor-propeller combinations achieve $80-85\,\%$ motor efficiency near maximum power and around 75\,\% at typical operating conditions, a constant motor efficiency of $\eta_M = 0.75$ is used, unless otherwise indicated. Only when highly accurate range and endurance estimates are required, the full motor model (\ref{eq:eta}) is used. In such cases, a thrust test stand is needed to identify the model coefficients for the given motor-propeller pairing. \review{Because the BEM model takes the full flow-state around the propeller into account measurements obtained on a static thrust test-stand transfer well to dynamic flights.}

\vspace*{0pt}
\section{Battery Model}
\vspace*{0pt}
This section explains the battery model developed in this work. Having an accurate model is a key component for precise range and endurance estimates. After all, the battery capacity is the limiting factor of the flight time. 

Two types of battery models could be used: Peukert models (battery capacity models) calculate the effective capacity of the battery when it is discharged at a fixed, given rate. Battery voltage models on the other hand estimate the terminal voltage of the LiPo battery given its state-of-charge (SoC) and the momentary power consumption. This work relies on the latter type of model, because it is more flexible as it can also be used in cases where the multicopter has non-constant power demand (e.g. battery-aware path planning for complex missions). 

\vspace*{-9pt}
\subsection{Model Structure}
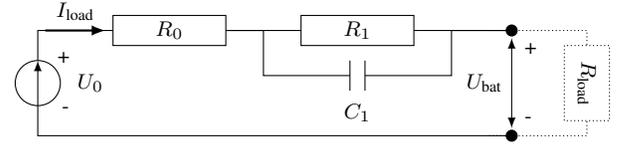
\begin{figure}
\centering
\begin{tikzpicture}[>=latex]
\tikzstyle{every node}=[font=\footnotesize]
\draw (0,0) circle (0.3cm) node [right = 4mm] {$U_0$} node [above right=1.4mm] {+} node [below right=2mm] {-};
\draw [->] (0,-0.3) -- ++ (0, 0.6);
\draw (1, 0.5) rectangle node {$R_0$} ++ (1.5,0.4) ;
\draw (3.5, 0.5) rectangle node {$R_1$} ++ (1.5,0.4) ;
\draw (4.15, -0.1) -- ++ (0, 0.4);
\draw (4.35, -0.1) -- ++ (0, 0.4);
\node at (4.25, -0.4) {$C_1$};
\draw [densely dotted] (7, 0.5) rectangle  node [rotate=-90] {$R_\text{load}$} ++ (0.6, -1);
\fill (6.3, 0.7) circle (0.08) node [below right=0.5mm] {+};
\fill (6.3, -0.7) circle (0.08) node [above right=0.5mm] {-};

\draw (0,0.3) |- (1,0.7);
\draw (2.5,0.7) -- (3.5, 0.7);
\draw (2.5,0.7) -- ++ (0.5, 0) |- (4.15, 0.1);
\draw (6.3, 0.7) -- ++ (-0.8, 0) |- (4.35, 0.1);
\draw (6.3, 0.7) -- ++ (-0.5, 0) |- (5, 0.7);
\draw (6.3, -0.7) -| (0, -0.3);
\draw [densely dotted] (6,0.7) -| (7.3, 0.5);
\draw [densely dotted] (6,-0.7) -| (7.3, -0.5);
\draw [<->] (6.3, -0.6) -- node [midway, left] {$U_\text{bat}$} (6.3,0.6);

\draw [->, thick] (0.1,0.7) -- (0.9, 0.7) node [midway, above] {$I_\text{load}$};

\end{tikzpicture}
\vspace*{-6pt}
    \caption{Thevenin equivalent circuit for the one-time-constant (OTC) battery model. The load does not need to be static but could, for example, be a multirotor aerial vehicle.}
    \label{fig:otc_circuit}
    \vspace*{-12pt}
\end{figure}

\vspace*{-2pt}
Battery voltage models leverage Thevenin equivalent circuits to predict the battery voltage. Fig.~\ref{fig:otc_circuit} shows the equivalent circuit diagram for the used OTC (one time constant) battery model. The voltage of the voltage source $U_0$ corresponds to the open-circuit voltage of the battery. When a possibly time-varying load is connected to the circuit and a current $I_\text{load}(t)$ flows, the voltage $U_\text{bat}(t)$ at the output terminals can be calculated as~\cite{2012:Rahmoun}
\begin{align}
\dot{U}_\text{cap}(t) &= \frac{-U_\text{cap}(t)}{R_1 \cdot C_1} + \frac{I_\text{load}(t)}{C_1} \,,\label{eq:u_cap}\\
U_\text{bat}(t) &= U_0(t) - U_\text{cap}(t) - R_0(t) I_\text{load}(t) \,,
\label{eq:Ubat_implicit}
\end{align}
where $R_0(t)$, $R_1$, $C_1$ are defined as shown in Fig~\ref{fig:otc_circuit}. 

When the load is a multicopter, only the power demand of the motors can be computed. Replacing the unknown $I_\text{load}(t)$ in (\ref{eq:Ubat_implicit}) with $P_\text{cell}(t)/U_\text{bat}(t)$ yields a quadratic equation in $U_\text{bat}$. Solving for the battery voltage gives the final result (the dependence on $t$ has been omitted for improved readability):
\begin{equation}
    U_\text{bat} = \frac{1}{2}\left(U_0 - U_\text{cap} - \sqrt{\left(U_0 - U_\text{cap}\right) - 4 R_0 P_\text{cell}}\right)\,.
    \label{eq:u_bat_final}
\end{equation}

To avoid coupling (\ref{eq:u_cap}) and (\ref{eq:Ubat_implicit}) the term $I_\text{load}(t)$ in (\ref{eq:u_cap}) is approximated by $k P_\text{cell}$ for some constant $k$. Finally, (\ref{eq:u_cap}) is reformulated as a lumped parameter model with time-constant $\tau_{RC}$ to yield:
\vspace*{-6pt}
\begin{equation}
    \dot{U}_\text{cap}(t) = \frac{k P_\text{cell} -U_\text{cap}(t)}{\tau_{RC}}\,.
    \label{eq:u_cap_final}
\end{equation}

Commercial LiPo batteries are available in different configurations. For example, a '4S' battery consists of ${N_\text{S}=4}$ LiPo cells in series and a '6S2P' battery is made of ${N_\text{S}=6}$ networks of which ${N_\text{P}=2}$ are connected in parallel. Furthermore, the batteries come with different capacities $C_\text{bat}$.

The goal is to develop a widely applicable range and endurance model for multicopters and hence the battery model must also be applicable to the myriad of available battery configurations. The development of the unified model is enabled by the normalization of the power $P_\text{cell}$ and consumed energy $E_\text{cell}$ to a single cell of the battery pack:
\begin{align}
P_\text{cell}(t) = \frac{P_\text{mot}(t)}{N_\text{cell} \cdot C_\text{cell}},\quad &&
E_\text{cell}(t) = \int_0^t P_\text{cell}(\tau)\,\mathrm{d}\tau,
\label{eq:normalized_power}
\end{align}

with $P_\text{mot}(t)$ the instantaneous power consumption of the multirotor, $N_\text{cell} = N_\text{S}N_\text{P}$ the total number of cells of the battery and $C_\text{cell} = C_\text{bat}/N_\text{P}$ the capacity per battery cell. With this, the average power consumption $\bar{P}_\text{cell}(t)$ of the whole flight can be written as 
\[\bar{P}_\text{cell}(t) = \frac{1}{t} \int_0^t P_\text{cell}(\tau)\,\mathrm{d}\tau = \frac{E_\text{cell}(t)}{t} \,.\]

Based on physical insights, the open-circuit voltage $U_0$ can only be a function of the state of charge (SoC), or equivalently, the amount of energy $E$ already consumed since the battery was fully charged. By minimizing the RMSE on held-out validation data, a third order polynomial~\cite{2018:Zhang} for the open-circuit battery voltage is identified:
\begin{equation}
    U_0(E_\text{cell}) = a_0 + a_1 E_\text{cell} + a_2 E_\text{cell}^2 + a_3 E_\text{cell}^3\,.
\end{equation}

The internal resistance of a LiPo battery strongly depends on the temperature. Unfortunately, this information is typically not available. The experiments show that the average power consumption is a good proxy since batteries heat up quickly when the power demand is high. Furthermore, the experiments show a strong dependency on the capacity of the cell. Thus, the following model is used:
\begin{equation}
    R_0(\bar{P}_\text{cell}, C_\text{cell}) = \text{max}(b_0 + b_1 \bar{P}_\text{cell} + b_2 C_\text{cell}, R_\text{min}) \,.
    \label{eq:internal_resistance}
\end{equation}

\vspace*{-16pt}
\subsection{Parameter Identification}
\vspace*{0pt}
To identify all parameters of the battery model, 10 different batteries ranging from '4S1P 1.55Ah' to '6S4P 5.2Ah' are tested with different, step-wise constant discharge profiles. In total, about \SI{5000}{\second} of battery discharge data is recorded. The discharge rates range from 5\,C to 70\,C.

The model parameters are identified using a two-step approach: first, the model parameters in (\ref{eq:internal_resistance}) are estimated from the steps in power consumption, as proposed by~\cite{2012:Rahmoun}. Subsequently, all other parameters ($a_{[0-3]}$, $k$, $\tau_{RC}$) are estimated simultaneously by numerically minimizing the RMSE between the model predictions and the real-world data from the battery tests. All model parameter estimates are summarized in Tab.~\ref{tab:values}. 

\begin{table}
\caption{\textnormal{Numerical values for the battery model coefficients.}}
\vspace*{-6pt}
\label{tab:values}
\setlength{\tabcolsep}{3pt}
\begin{tabularx}{1\linewidth}{>$l<$p{1cm} >$l<$X >$l<$X >$l<$X}
\toprule
a_0 & \SI{4.2}{\volt}  & a_1 &  -0.1102178 & a_2 &  0.0103368 & a_3 & -4.3778e-4  \\
\midrule
R_\text{min} & \SI{4.5}{\milli\ohm} & b_0 & 0.0015778  & b_1 &  -7.7608e-5 & b_2 & 0.0069498 
  \\
\midrule
\tau_{RC} & \SI{3.3}{\second} &  k & 0.00104846 \\
\bottomrule
\end{tabularx}
\vspace*{-12pt}
\end{table}

\begin{figure}[t]
\vspace*{2pt}
    \centering
    \tikzstyle{every node}=[font=\footnotesize]
\definecolor{mycolor1}{rgb}{0.00000,0.44700,0.74100}%
\definecolor{mycolor2}{rgb}{0.85000,0.32500,0.09800}%
\definecolor{mycolor3}{rgb}{0.92900,0.69400,0.12500}%
\definecolor{mycolor4}{rgb}{0.49400,0.18400,0.55600}%
\begin{tikzpicture}

\begin{axis}[%
width=3.244cm,
height=4cm,
at={(0cm,0cm)},
scale only axis,
xmode=log,
xmin=1.0000,
xmax=1000.0000,
xminorticks=true,
xlabel={Time [s]},
ymin=14.0000,
ymax=17.0000,
ylabel={4S Battery Voltage [V]},
axis background/.style={fill=white},
xmajorgrids,
xminorgrids,
ymajorgrids,
legend style={legend cell align=left, align=left, draw=white!15!black},
legend columns=1,
xlabel shift = -2pt,
ylabel shift = -6pt,
line join = round,
scale only axis=true,
legend pos= north east,
height = 2.5cm,
legend columns=2,
legend style={at={(axis cs:1,17.1)}, anchor=south west},
]
\addplot [color=mycolor1]
  table[row sep=crcr]{%
0.9750	16.5612\\
3.2500	16.5472\\
6.7500	16.5258\\
11.5000	16.4971\\
17.6000	16.4608\\
25.2000	16.4163\\
34.4750	16.3633\\
45.7750	16.3004\\
59.5250	16.2264\\
76.4250	16.1391\\
97.4750	16.0360\\
124.3500	15.9130\\
159.9500	15.7638\\
210.6500	15.5764\\
297.3500	15.3141\\
531.9250	14.8405\\
637.6750	14.6668\\
721.7500	14.5141\\
800.2250	14.3436\\
878.5000	14.1326\\
919.8500	14.0000\\
};
\addlegendentry{100 W}

\addplot [color=mycolor2]
  table[row sep=crcr]{%
0.9750	16.3472\\
2.4000	16.3295\\
4.4250	16.3047\\
7.1000	16.2723\\
10.4750	16.2320\\
14.6250	16.1834\\
19.6750	16.1258\\
25.8250	16.0576\\
33.3250	15.9775\\
42.6000	15.8829\\
54.2500	15.7708\\
69.3250	15.6362\\
89.8000	15.4711\\
120.5750	15.2574\\
185.4250	14.9148\\
286.7500	14.5456\\
333.9250	14.3854\\
374.4750	14.2278\\
413.4250	14.0438\\
421.5750	13.9999\\
};
\addlegendentry{200 W}

\addplot [color=mycolor3]
  table[row sep=crcr]{%
0.9750	16.0001\\
1.9000	15.9770\\
3.1500	15.9462\\
4.7250	15.9078\\
6.6750	15.8613\\
9.0500	15.8058\\
11.9250	15.7405\\
15.4250	15.6637\\
19.7250	15.5732\\
25.1000	15.4660\\
31.9750	15.3380\\
41.1250	15.1823\\
54.2500	14.9861\\
77.3500	14.7062\\
135.7250	14.2323\\
161.1750	14.0592\\
169.3750	13.9999\\
};
\addlegendentry{400 W}

\addplot [color=mycolor4]
  table[row sep=crcr]{%
0.9750	15.6667\\
1.6000	15.6356\\
2.3750	15.5977\\
3.3500	15.5509\\
4.5250	15.4957\\
5.9500	15.4306\\
7.7000	15.3533\\
9.8500	15.2622\\
12.5250	15.1548\\
15.9500	15.0263\\
20.5000	14.8704\\
27.0250	14.6735\\
38.4250	14.3943\\
61.4750	13.9998\\
};
\addlegendentry{800 W}

\end{axis}

\begin{axis}[%
width=3.244cm,
height=2.5cm,
at={(4.268cm,0cm)},
scale only axis,
xmin=0.0000,
xmax=600.0000,
xlabel={Power [W]},
ymin=19.0000,
ymax=27.0000,
ylabel={Effective Cap. [Wh]},
axis background/.style={fill=white},
xmajorgrids,
ymajorgrids,
legend style={legend cell align=left, align=left, draw=white!15!black},
legend columns=2,
xlabel shift = -2pt,
ylabel shift = -6pt,
line join = round,
scale only axis=true,
legend pos= north east,
legend style={at={(axis cs:600,27.26)}, anchor=south east}
]
\addplot[only marks, mark=*, mark options={}, mark size=1.0000pt, color=mycolor1, fill=mycolor1] table[row sep=crcr]{%
x	y\\
245.5357	24.1003\\
389.3150	22.9571\\
501.7457	20.3830\\
555.8564	19.6876\\
139.3083	24.2522\\
637.8149	20.5208\\
123.2657	24.0339\\
526.4765	20.4045\\
522.0681	20.7725\\
600.9918	20.0414\\
492.6112	21.8576\\
395.6778	21.7097\\
376.0539	22.1816\\
301.7873	23.1746\\
608.1907	20.5101\\
181.9782	24.7363\\
279.5164	23.0101\\
365.5480	22.2484\\
435.4608	21.7201\\
520.2363	19.9152\\
533.1960	20.6808\\
477.3411	22.2647\\
511.3521	21.0211\\
546.7603	20.6033\\
492.5204	21.3663\\
81.6556	25.2449\\
51.8380	26.1445\\
174.1618	22.9517\\
200.0246	23.9188\\
182.1505	24.1877\\
202.8302	23.3824\\
189.9817	23.5614\\
95.9844	25.1689\\
32.0475	26.2098\\
};
\addlegendentry{Meas.}

\addplot [color=mycolor2]
  table[row sep=crcr]{%
1.0000	26.6400\\
51.0000	25.9635\\
104.0000	25.2360\\
198.0000	23.9735\\
270.0000	23.0421\\
314.0000	22.4987\\
374.0000	21.8042\\
418.0000	21.3381\\
456.0000	20.9713\\
507.0000	20.5408\\
548.0000	20.2521\\
586.0000	20.0351\\
601.0000	19.9635\\
};
\addlegendentry{Ours}

\addplot [color=mycolor3]
  table[row sep=crcr]{%
1.0000	26.6076\\
6.0000	26.4878\\
16.0000	26.2868\\
32.0000	26.0032\\
54.0000	25.6517\\
81.0000	25.2575\\
113.0000	24.8270\\
150.0000	24.3660\\
191.0000	23.8913\\
237.0000	23.3954\\
287.0000	22.8927\\
341.0000	22.3857\\
400.0000	21.8681\\
463.0000	21.3517\\
530.0000	20.8381\\
601.0000	20.3289\\
};
\addlegendentry{Gen. Peuk.}

\addplot [color=mycolor4]
  table[row sep=crcr]{%
24.0000	27.0199\\
27.0000	26.7766\\
31.0000	26.4941\\
35.0000	26.2483\\
40.0000	25.9805\\
45.0000	25.7465\\
51.0000	25.5002\\
58.0000	25.2496\\
66.0000	25.0003\\
75.0000	24.7561\\
85.0000	24.5192\\
97.0000	24.2718\\
111.0000	24.0218\\
127.0000	23.7747\\
145.0000	23.5339\\
166.0000	23.2907\\
190.0000	23.0504\\
217.0000	22.8164\\
248.0000	22.5836\\
284.0000	22.3498\\
325.0000	22.1195\\
373.0000	21.8867\\
428.0000	21.6568\\
491.0000	21.4296\\
564.0000	21.2027\\
601.0000	21.0995\\
};
\addlegendentry{Peuk.}

\end{axis}

\end{tikzpicture}%
\tikzstyle{every node}=[font=\normalsize]
    \vspace*{-20pt}
    \caption{The left plot shows discharge curves for a typical 4S 1.8Ah battery. A higher power demand leads to lower voltages as well as a much shorter flight time. The right plot shows \review{how well different models predict the Peukert effect: higher discharge rates lead to a reduction in effective capacity. The proposed model ("Ours") is compared with the original Peukert model ("Peuk.") and the state-of-the-art generalized Peukert model ("Gen. Peuk."). It can be seen that, albeit being a voltage model, the proposed model fits the measurements well.}}
    \label{fig:battery_model}
    \vspace*{-10pt}
\end{figure}
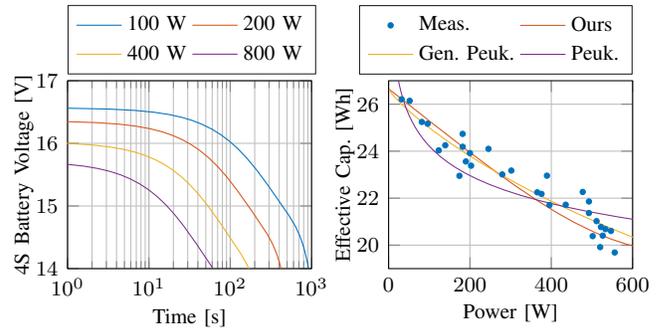

The identified battery model is very accurate and achieves a relative error as low as 1.3\,\% (\SI{43.1}{\milli\volt}) RMSE across all real-world experiments with highly varying power demand. 
\review{
In a constant-power discharge setting, the proposed model can be compared to capacity battery models. 
The left plot in Fig.~\ref{fig:battery_model} utilizes the proposed model to show how the battery voltages decreases over time. 
The effective capacity can be calculated by integrating the power up to the time where the battery is discharged to a cell voltage of \SI{3.5}{\volt}. 
The right plot shows that the effective capacity of the battery changes as a function of the power draw.
Results obtained with the original Peukert model \cite{1897:Peukert}, a state-of-the-art generalized Peukert model \cite{2020:Galushkin}, and the proposed model are compared with experimental data obtained in a constant discharge power setting. 
The Peukert capacity model and generalized Peukert capacity model achieve a RMSE of \SI{0.82}{\watt\hour} and \SI{0.53}{\watt\hour}, respectively. 
The proposed battery voltage model\textemdash only implicitly containing information about the battery capacity\textemdash nearly reaches the accuracy of the state-of-the-art capacity model (RMSE  \SI{0.58}{\watt\hour}). 
Because of its applicability to general, non-constant power flight profiles, the proposed battery voltage model is used subsequently. Only for the special case of straight, constant velocity flights a Peukert model could be used.}

\vspace*{-6pt}
\section{Validation: Real-World Experiments}
\vspace*{0pt}
This section presents the validation of the proposed approach to range, endurance, and optimal flight speed estimation by comparing its predictions with real-world experiments. 

To collect the flight data needed for validation, experiments inside a flying arena are conducted. It is equipped with a motion capture system (tracking volume: $\SI{25}{\meter}\times\SI{25}{\meter}\times\SI{8}{\meter}$) used for state estimation and control. The experimental platform is a custom built quadrotor based on a 6\,inch frame. It weighs \SI{752}{\gram} and is equipped with 2400\,KV HobbyWing XRotor 2306 motors and 5.1\,inch Azure 5148 propellers. Batteries of type \SI{1.8}{\ampere\hour} 4S 120\,C are used.

\vspace*{-10pt}
\subsection{Power Consumption \& Optimal Speed}
\vspace*{-2pt}
In the first set of experiments, the multicopter simulator is used to fly a circle-trajectory ($r=5\,m$) in simulation and all parameters (e.g. axial propeller torque, motor speeds) are recorded. This information is used by the detailed motor efficiency model (\ref{eq:eta}) to calculate the power demand of the whole multicopter. Then the same trajectory is also flown in the real world and the power consumption is logged aboard the vehicle. \review{The parameters of the simulation are not tuned using the real-world flight data as they are identified purely from static thrust test-stand measurements as discussed above.}

\begin{figure}[t]
    \centering
    \tikzstyle{every node}=[font=\footnotesize]
\definecolor{mycolor1}{rgb}{0.00000,0.44700,0.74100}%
\definecolor{mycolor2}{rgb}{0.85000,0.32500,0.09800}%
\begin{tikzpicture}

\begin{axis}[%
width=3.244cm,
height=2.9cm,
at={(0cm,0cm)},
scale only axis,
xmin=0.0000,
xmax=10.5000,
xlabel={Flight Speed [m/s]},
ymin=0.0000,
ymax=400.0000,
ylabel={Power [W]},
axis background/.style={fill=white},
xmajorgrids,
ymajorgrids,
legend style={legend cell align=left, align=left, draw=white!15!black, inner sep=2pt},
legend columns=2,
xlabel shift = -2pt,
ylabel shift = -3pt,
line join = round,
scale only axis=true,
legend pos=south east
]
\addplot [color=mycolor1, only marks, mark=asterisk, mark options={solid, mycolor1}]
  table[row sep=crcr]{%
4.0000	142.0000\\
6.0000	175.0000\\
8.0000	251.0000\\
10.0000	377.0000\\
};
\addlegendentry{Meas.}

\addplot [color=mycolor2]
  table[row sep=crcr]{%
0.1000	134.1228\\
0.6200	134.0689\\
1.8400	133.8673\\
2.0600	133.9378\\
2.2900	134.0804\\
2.5300	134.3380\\
2.8000	134.7986\\
3.0000	135.2758\\
3.2000	135.8911\\
3.4200	136.7552\\
3.6900	138.1322\\
4.0000	140.1989\\
4.3100	142.8808\\
4.6300	146.3790\\
4.9400	150.5707\\
5.2700	156.0016\\
5.6200	162.9847\\
5.9500	170.7835\\
6.2200	178.1403\\
6.6500	191.8326\\
7.1300	210.0830\\
7.5700	229.6167\\
7.9900	250.9939\\
8.4200	275.8361\\
8.9500	310.4687\\
9.4700	349.1283\\
10.0800	400.5753\\
10.5100	440.8404\\
};
\addlegendentry{Sim.}

\end{axis}

\begin{axis}[%
width=3.244cm,
height=2.9cm,
at={(4.268cm,0cm)},
scale only axis,
xmin=0.0000,
xmax=10.5000,
xlabel={Flight Speed [m/s]},
ymin=0.0000,
ymax=150.0000,
ylabel={Range per Wh [m]},
axis background/.style={fill=white},
xmajorgrids,
ymajorgrids,
legend style={legend cell align=left, align=left, draw=white!15!black},
legend columns=2,
xlabel shift = -2pt,
ylabel shift = -6pt,
line join = round,
scale only axis=true,
legend pos=south east
]
\addplot [color=mycolor1, only marks, mark=asterisk, mark options={solid, mycolor1}]
  table[row sep=crcr]{%
4.0000	101.4085\\
6.0000	123.4286\\
8.0000	114.7410\\
10.0000	95.4907\\
};
\addlegendentry{Meas.}

\addplot [color=mycolor2]
  table[row sep=crcr]{%
0.1000	2.6841\\
3.1200	82.8178\\
3.7400	97.2625\\
4.2100	106.7746\\
4.6100	113.5635\\
4.9500	118.2325\\
5.2400	121.3391\\
5.4800	123.2773\\
5.6800	124.4568\\
5.8700	125.1971\\
6.0300	125.5661\\
6.0800	125.5735\\
6.1500	125.6972\\
6.2800	125.6675\\
6.4000	125.5032\\
6.6100	124.9627\\
6.8100	124.0926\\
7.0400	122.7902\\
7.3000	120.9585\\
7.6700	117.7621\\
8.1900	112.4547\\
10.5100	85.8270\\
};
\addlegendentry{Sim.}

\end{axis}
\end{tikzpicture}%
\tikzstyle{every node}=[font=\normalsize]
    \vspace*{-21pt}
    \caption{The plots show a comparison between real-world experiments and pure simulation results. On the left, the vehicle's power consumption \review{on a circular trajectory} is plotted as a function of the flight speed. The right plot shows the energy consumption per distance covered as a function of the flight speed. Simulation and experiment match very well, validating the accuracy of the multicopter simulator and motor model.}
    \label{fig:circle_flights}
    \vspace*{0pt}
\end{figure}
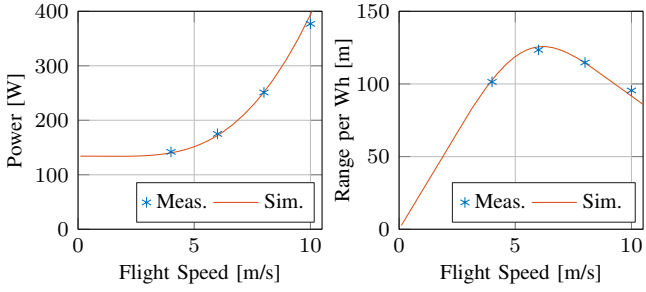

Fig.~\ref{fig:circle_flights} shows the results of this experiment: the measured power consumption and the measured specific energy consumption (energy consumed per distance covered) match their simulated counterparts extremely well, with 2.5\% RMSE in power consumption. Note that the minimum of the simulated specific energy consumption also aligns with the measurements. This means, that the simulation can be used to accurately calculate the the optimal flight speed to achieve a maximum flight range. 

\vspace*{-9pt}
\subsection{Battery Voltage}
\vspace*{-2pt}
For the second set of experiments, the battery model is also added to the simulation, as the simulated power demand of the multicopter is fed into the battery model which predicts the battery voltage. Similar to the first set of experiments, flights conducted in simulation are repeated in the real world.

\begin{figure}[t]
    \centering
    \input{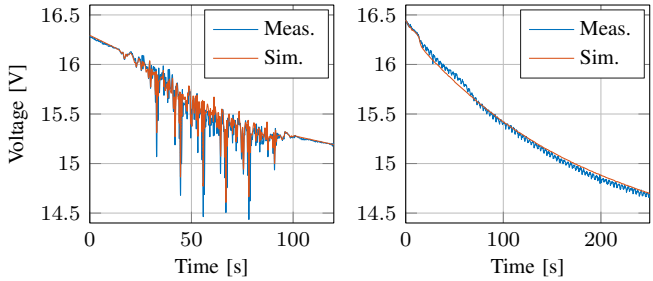}
    \vspace*{-22pt}
    \caption{Comparison between the simulated battery voltage and the measured one. The \review{left plot shows the simulation accuracy} while tracking an aggressive trajectory with large variations in power consumption. The \review{right plot shows the battery voltage} when the vehicle was flying in circles for over 4\,min. The small oscillations observed in the real experiment are due to the controller not tracking the circle trajectory perfectly. It can be seen that simulation and experiment match very well.}
    \label{fig:closed_loop_simulation}
    \vspace*{-10pt}
\end{figure}

Fig.~\ref{fig:closed_loop_simulation} shows a comparison of simulated and measured battery voltages during two different flights. The first flight is quite challenging to model, because the multicopter follows a very agile trajectory which causes large variations in instantaneous power demand. The second trajectory is a circle flown at a constant speed of \SI{6}{\meter\per\second}. As can be seen in Fig.~\ref{fig:closed_loop_simulation}, the predicted battery voltage matches the experiments very well, with an RMSE of only \SI{60.8}{\milli\volt}.  This provides strong evidence for that not only the battery model is accurate, but that it can be used together with the BEM simulator and the motor model for precise range and endurance estimation.

\section{Range, Endurance and Optimal Speed}
\vspace*{0pt}
The models presented and validated in the last sections, are of limited use if a simple range, endurance or optimal speed estimate is required. To use the approach presented so far one would for example need to implement a multicopter simulator. This section presents a more accessible approach, by limiting the task to range, endurance, and optimal flight speed estimation for straight-line flight. The full-fledged simulation is used to simulate dozens of different multicopters and based on the simulation results, a simple empirical model is developed.

\vspace*{-8pt}
\subsection{Power Consumption}
\vspace*{-1pt}
The decisive factor limiting the range and endurance is the power consumption of the multicopter. At hover, the power demand can be calculated from momentum-theory using (\ref{eq:hover_power}).  The simulation results suggest that a constant ratio between the hover power $P_h$ and the power consumption at the operating points maximizing the endurance $P_e$ or range $P_r$ may be a sufficient approximation. We find that
\begin{equation}
    \frac{P_r}{P_h} = 1.092 \pm [0.0361], \qquad \frac{P_e}{P_h} = 0.914 \pm [0.0323]
    \label{eq:power_ratios}
\end{equation}
where the number given in brackets denotes the standard deviation. Note that $P_e/P_h < 1$ means that flying forward at slow speeds reduces the power consumption compared to hovering and hence increases the endurance. This is a known phenomenon~\cite{2017:Karydis} and can be explained using dynamic lift: the required propeller speed reduces because the propeller generates more lift.
\review{The power consumption of the flight controller and potentially the onboard computer is neglected as it is typically negligible compared to the power consumption of the motors. For example, an NVIDIA Jetson TX2 consumes \SI{<10}{\watt} while the research platform needs \SI{130}{\watt} to hover.}

\vspace*{-9pt}
\subsection{Optimal Speed}
\vspace*{-1pt}
Simulation experiments show that the optimal flight speed for maximal range $v_r$ or maximal endurance $v_e$ depend mostly on the induced velocity at hover and the surface area $A$ of the multicopter. To simplify the analysis, the normalized quantities $\hat v_e = v_e/v_{i,h}$ and $\hat v_r = v_r/v_{i,h}$ are introduced. \review{Based on the numerous conducted simulation experiments, a step-wise model selection procedure is carried out and it has} been found that a linear model of the form
\begin{equation}
    \hat v_{[e|r]}^{-1} = c_{0,r|e} + c_{1,r|e} v_{i,h} + c_{2,r|e} A
    \label{eq:v_opt_inv}
\end{equation}
fits well, with \review{a coefficient of determination} $R^2 = 0.97$ and $R^2 = 0.966$ for the endurance and range estimates, respectively. Equation (\ref{eq:v_opt_inv}) therefore allows one to directly calculate the optimal flight speeds for maximal endurance and maximal range. The numeric values for the fitted coefficients are given in Table~\ref{tab:coeffs}.

\vspace*{-8pt}
\review{\subsection{Wind}}
\vspace*{-1pt}
\review{To account for situations where a constant head or tailwind is present, a wind correction factor $k_{w,v}$ for the optimal range velocity $v_r$ is introduced. Similarly, a wind correction factor $k_{w,P}$ for the corresponding power consumption is needed. For strong tailwinds the vehicle is carried primarily by the wind and hence the optimal range velocity approaches the one for the optimal endurance. For strong headwinds, forward flight becomes increasingly power-intensive and it is best to fly only slightly faster than the oncoming headwind. Based on those preliminary considerations, a simple model is fitted based on the BEM experiments.}
\review{
\begin{align}
    k_{w,v} = \frac{v_{r,w}}{v_r} = \frac{\ln (1 + \exp(c_{0,w} (\frac{v_{w}}{v_r} - c_{1,w})))}{c_{0,w}} + c_{2,w} 
    \label{eq:wind_velocity} \\
    k_{w,P} = {P_{r,w}}/{P_{r}} = \exp(c_{0,P} \,v_{w}/v_r - c_{1,P}) + c_{2,P}
    \label{eq:wind_power}
\end{align}
where ${v}_{r,w}$, $P_{r,w}$ are the wind-corrected optimal range speed and power consumption. The wind velocity  $v_w$ is measured relative to the drone, e.g. $v_w < 0$ for tailwinds.
The numeric values for the coefficients are given in Table~\ref{tab:coeffs}.
}

\vspace*{-8pt}
\subsection{Battery Model}
\vspace*{-1pt}
From (\ref{eq:v_ind_h}), (\ref{eq:hover_power}), (\ref{eq:power_ratios}), and (\ref{eq:v_opt_inv}) the power consumption and the speed that lead to maximum range and endurance are known. To complete the modeling, only the decrease in effective battery capacity needs to be accounted for. 

Instead of using the full model, the relative capacity $\kappa$ can be approximated as a \review{third-order polynomial} function of the normalized power consumption $P_\text{cell}$ defined in (\ref{eq:normalized_power}):
\begin{equation}
    \kappa = C_\text{eff}/C = d_0 + d_1 P_\text{cell} + d_2 P_\text{cell}^2 + d_3 P_\text{cell}^3
    \label{eq:cubic_battery}
\end{equation}
The values for the coefficients are summarized in Table~\ref{tab:coeffs}.

\begin{table}[t!]
    \centering
    \caption{\textnormal{Numeric values for the fitted coefficients.}}
    \vspace*{-3pt}
    \def\arraystretch{1.1}
    \label{tab:coeffs}
    \begin{tabularx}{1\linewidth}{>$l<$X|>$l<$X|>$l<$X}
    \toprule
        \multicolumn{2}{c|}{Endurance} & \multicolumn{2}{c|}{Range} & \multicolumn{2}{c}{Battery} \\
        \midrule
         c_{0,e} & 0.10188      &  c_{0,r} & 0.041546     &  d_0 & 0.9876     \\
         c_{1,e} & 0.071358     &  c_{1,r} & 0.041122     &  d_1 & -0.0020    \\
         c_{2,e} & 0.0007381    &  c_{2,r} & 0.00053292   &  d_2 & -5.2484e-05 \\
                 &              &          &              &  d_3 & 1.2230e-07\\
        \midrule \midrule
        \multicolumn{2}{c|}{Wind, Speed} & \multicolumn{2}{c|}{Wind, Power} \\
        \midrule
        c_{0,w} & 1.5730        &   c_{0,P} & 2.4000  \\
        c_{1,w} & 0.5477        &   c_{1,P} & 2.0998  \\ 
        c_{2,w} & 0.7732        &   c_{2,P} & 0.8763  \\
        \bottomrule    
    \end{tabularx}
\vspace*{-12pt}
\end{table}

\vspace*{-8pt}
\subsection{Algorithm and Example}
\vspace*{-1pt}
\begin{table*}[t]
\review{
    \centering
    \caption{\textnormal{A comparison of no-wind range and endurance estimates for various commercially available drones. The pen-and-paper algorithm gives very good estimates.}}
    \label{tab:range_commercial}
    \begin{tabularx}{1\linewidth}{X|c@{\hskip 8pt}l@{\hskip 8pt}r@{\hskip 4pt}r@{\hskip 8pt}r|l@{\hskip 8pt}l|l@{\hskip 8pt}r}
    \toprule
    Drone Model & \multicolumn{5}{c|}{Physical Parameters from Manufacturer} & \multicolumn{2}{c|}{Endurance} & \multicolumn{2}{c}{Range} \\[3pt]
    & \multicolumn{1}{c}{Mass}   
    & \multicolumn{1}{c}{Propeller}  
    & \multicolumn{2}{c}{LiPo Battery} 
    & \multicolumn{1}{c|}{Area}  
    & \multicolumn{1}{c}{Spec.} 
    & \multicolumn{1}{c|}{\textbf{Ours}} 
    & \multicolumn{1}{c}{Specification} 
    & \multicolumn{1}{c}{\textbf{Ours}} \\
    \midrule
    DJI Mavic 2 
        & \SI{0.91}{\kilo\gram}
        & 4 $\times$ \SI{11.0}{\centi\meter} 
        & 4S1P 
        & \SI{3.9}{\ampere\hour} 
        & \SI{200}{\centi\meter\squared} 
        & \SI{31}{\minute} 
        & \SI{33}{\minute} 
        & \SI{18}{\kilo\meter} @ \SI{50}{\kilo\meter\per\hour} 
        & \SI{24}{\kilo\meter} @ \SI{51}{\kilo\meter\per\hour} \\
        DJI Mavic 3 
        & \SI{0.90}{\kilo\gram}
        & 4 $\times$ \SI{11.9}{\centi\meter} 
        & 4S1P 
        & \SI{5.0}{\ampere\hour} 
        & \SI{215}{\centi\meter\squared} 
        & \SI{46}{\minute} 
        & \SI{48}{\minute} 
        & \SI{30}{\kilo\meter}
        & \SI{32}{\kilo\meter} @ \SI{48}{\kilo\meter\per\hour} \\
    DJI Matrice 200
        & \SI{6.14}{\kilo\gram}
        & 4 $\times$ \SI{21.6}{\centi\meter} 
        & 6S2P 
        & \SI{15.3}{\ampere\hour} 
        & \SI{1700}{\centi\meter\squared} 
        & \SI{24}{\minute} 
        & \SI{23}{\minute} 
        & --- 
        & \SI{6.2}{\kilo\meter} @ \SI{19}{\kilo\meter\per\hour} \\
    DJI Matrice 600 Pro
        & \SI{15.5}{\kilo\gram}
        & 6 $\times$ \SI{26.7}{\centi\meter} 
        & 6S6P 
        & \SI{34.2}{\ampere\hour} 
        & \SI{1760}{\centi\meter\squared} 
        & \SI{18}{\minute} 
        & \SI{18}{\minute} 
        & --- 
        & \SI{5.4}{\kilo\meter} @ \SI{20}{\kilo\meter\per\hour} \\
    Parrot Anafi AI
        & \SI{0.90}{\kilo\gram}
        & 4 $\times$ \SI{5.7}{\centi\meter} 
        & 4S1P 
        & \SI{6.8}{\ampere\hour} 
        & \SI{400}{\centi\meter\squared} 
        & \SI{32}{\minute} 
        & \SI{31}{\minute} 
        & \SI{23}{\kilo\meter} @ \SI{50}{\kilo\meter\per\hour} 
        & \SI{23}{\kilo\meter} @ \SI{53}{\kilo\meter\per\hour} \\
    Skydio 2
        & \SI{0.78}{\kilo\gram}
        & 4 $\times$ \SI{8.5}{\centi\meter} 
        & 3S1P 
        & \SI{4.3}{\ampere\hour} 
        & \SI{268}{\centi\meter\squared} 
        & \SI{23}{\minute} 
        & \SI{26}{\minute} 
        & ---
        & \SI{18}{\kilo\meter} @ \SI{49}{\kilo\meter\per\hour} \\
    \end{tabularx}
    \vspace*{-6pt}
    }
\end{table*}
The full algorithm to calculate the range, endurance, and optimal speed for multicopters is summarized below. \review{This algorithm has been applied to six different commercially available drones. The specifications of the vehicles and the results are summarized in Table~\ref{tab:range_commercial}. The calculated performances closely match the manufacturers' specifications. As an example, let us consider the DJI Mavic 3 Quadcopter.
}

\begin{enumerate}
\setlength{\abovedisplayskip}{5pt}
\setlength{\belowdisplayskip}{5pt}
\setlength{\abovedisplayshortskip}{4pt}
\setlength{\belowdisplayshortskip}{4pt}
    \item Calculate the induced velocity at hover using (\ref{eq:v_ind_h}) and the power consumption at hover using (\ref{eq:hover_power}). For the DJI drone, we get
    \[ \review{ v_{i,h} = \SI{4.51}{\meter\per\second}, \qquad P_h = \SI{73.5}{\watt}\,.} \]
    \item Based on the hover power, the power consumption at the operating points for optimal endurance and optimal range can be calculated using (\ref{eq:power_ratios}). \review{If wind is considered, additionally use (\ref{eq:wind_power}).} We get
    \[ \review{P_{e} = \SI{67.2}{\watt}, \qquad P_r = \SI{80.2}{\watt}\,. }\]
    \item The motor model (\ref{eq:motor_model}) with $\eta_M \equiv 0.75$ is used to get the electric power demand
    \[ \review{ P_{\text{mot},e} =\SI{89.5}{\watt}, \qquad P_{\text{mot},r} = \SI{107.0}{\watt}\,.}\]
    \item From this, the normalized per cell power consumption can be calculated using (\ref{eq:normalized_power}). We get
    \[ \review{ P_{\text{cell},e} = \SI{4.48}{\ampere\hour\per\watt}, \qquad P_{\text{cell},r} = \SI{5.35}{\ampere\hour\per\watt}\,.}\]
    \item Now the simplified battery model (\ref{eq:cubic_battery}) can be employed to calculate the effective battery capacity. For the considered example, the values are
    \[ \review{ C_{\text{eff},e} = \SI{4.89}{\ampere\hour}, \qquad C_{\text{eff},r} = \SI{4.88}{\ampere\hour}\,.}\]
    \item From this, the maximum endurance $t_e$ and of the vehicle is readily obtained 
    \[ t_e = \frac{C_{\text{eff},e} \cdot \SI{3.7}{\volt} \cdot N_S \cdot \SI{3600}{\second}}{P_{\text{mot},e}} \,.\]
    The flight time $t_r$ at the maximum range operating point is calculated similarly. We get
    \[ \review{ t_e = \SI{2909}{\second}, \qquad  t_r = \SI{2429}{\second}\,. }\]
    \item To calculate the maximum range, the optimal flight speed needs to be computed using (\ref{eq:v_opt_inv}). \review{In case wind is considered, additionally use (\ref{eq:wind_velocity}).} We get
    \[ \review{ v_e = \frac{v_{i,h}}{\hat{v}^{-1}_e} = \SI{7.75}{\meter\per\second}, \quad v_r = \frac{v_{i,h}}{\hat{v}^{-1}_r} = \SI{13.12}{\meter\per\second}. }\]
    \item Finally, the maximum range $x_r$ can be calculated as:
    \[ \review{ x_r = t_r v_r = \SI{32.1}{\kilo\meter} }\]
\end{enumerate}

The endurance ${t_e = \SI{48}{\min}}$ calculated using the algorithm above matches the manufacturer's specification (\SI{46}{\min}) with an error less than 10\,\%. The range estimate ${x_r = \SI{48}{\kilo\meter}}$ also marginally exceeds the DJI specification (\SI{46}{\kilo\meter}). A possible reason is that the test flights conducted by DJI necessarily include a takeoff and landing phase which is neglected by the model or \review{that a more conservative battery safety threshold is used by DJI.}

\vspace*{-8pt}
\section{Conclusion}
\vspace*{-1pt}
This work presented a general and widely applicable approach to estimate the range, endurance, and optimal speed of multicopters. The method combines three models: a blade-element-momentum theory (BEM) based multicopter aerodynamics simulator, a motor model and a graybox battery model. The BEM model was validated with real-world flight data at speeds up to \SI{65}{\kilo\meter\per\hour} where it predicts the thrust force with only \SI{0.91}{\newton} RMSE. To account for the losses inside the electric motor, a model based on measurements of 44 different motor-propeller combinations was developed. The modeling was complemented with a graybox battery model identified from nearly \SI{2}{\hour} of measurement data gathered from 10 different battery configurations under different discharge profiles.

The combined model consisting of all three components was also verified through real-world experiments. The power consumption was calculated with an average accuracy of 2.5\,\%. The battery voltage model achieves a remarkably low error of only \SI{61}{\milli\volt} when compared to the experimental data. 

In addition to the highly accurate model, a simplified pen-and-paper algorithm was developed based on experiments leveraging the complete simulation.  It allows researchers, companies, and policy-makers alike to quickly and accurately estimate the characteristics of a given vehicle design. In the \review{six examples presented, the estimated range, endurance and the optimal flight are almost always within 10\,\% of the manufacturers' specifications}.

\bibliographystyle{ieeetr}
\bibliography{references}

\end{document}